# Pretraining and Distillation Matter More Than Architecture Family for Label-Free Single-Cell Classification

Philip Graemer[1,2] * and Giuseppe Di Caprio[1,2] *

[1]Centre for the Cellular Microenvironment, Department of Biomedical Engineering, University of Strathclyde, Glasgow, UK

[2]School of Molecular Biosciences, University of Glasgow, Glasgow, UK

* Corresponding author: {philip.graemer,giuseppe.dicaprio}@strath.ac.uk

**Abstract:** Choosing a deep learning architecture for label-free single-cell classification remains an open problem, with recent microscopy benchmarks reporting conflicting conclusions about the relative merits of convolutional and transformer-based models. We present a controlled benchmark on single-cell crops from LIVECell, a large public phase-contrast microscopy dataset, using source-image-disjoint train/validation/test splits to eliminate parent-image leakage and matched optimisation, augmentation, and evaluation protocols across EfficientNet, standard Vision Transformer (ViT), and EVA-02 models. This design enables the effects of architecture, pretraining, fine-tuning strategy, tokenisation, and distillation to be disentangled. Our results show that a previously reported CNN advantage is largely explained by pretraining rather than architecture alone; indeed, the smallest pretrained model outperforms the strongest trained from scratch despite far fewer parameters. Pretraining a fixed architecture improves macro-F1 by three to four points, whereas the gap between the best convolutional and best transformer models under pretrained initialisation is below half a point. Architectural choices remain consequential in more specific ways. In particular, ViT-S/8 outperforms ViT-S/16, and matches the four-times-larger ViT-B/16 at a quarter of the parameters, albeit at comparable compute, indicating that finer tokenisation is advantageous for small single-cell crops. We further find that layer-wise learning-rate decay, a key component of the original EVA-02 fine-tuning recipe, degrades performance in this setting, suggesting that transfer heuristics developed for natural-image recognition do not necessarily carry over to phase-contrast microscopy. Beyond backbone comparisons, we show that knowledge distillation offers a more effective route to improving the deployment frontier. Compact EfficientNet-B0 students distilled from teacher councils (averaging multiple models) surpass every individually trained backbone evaluated, including the EfficientNet-B5 and EVA-02 teacher backbones from which the councils were formed.

Together, these results demonstrate that rigorous control of pretraining and evaluation protocol is essential for interpreting biomedical architecture benchmarks, and that distillation may be a more impactful lever for practical single-cell classification than architecture family alone.

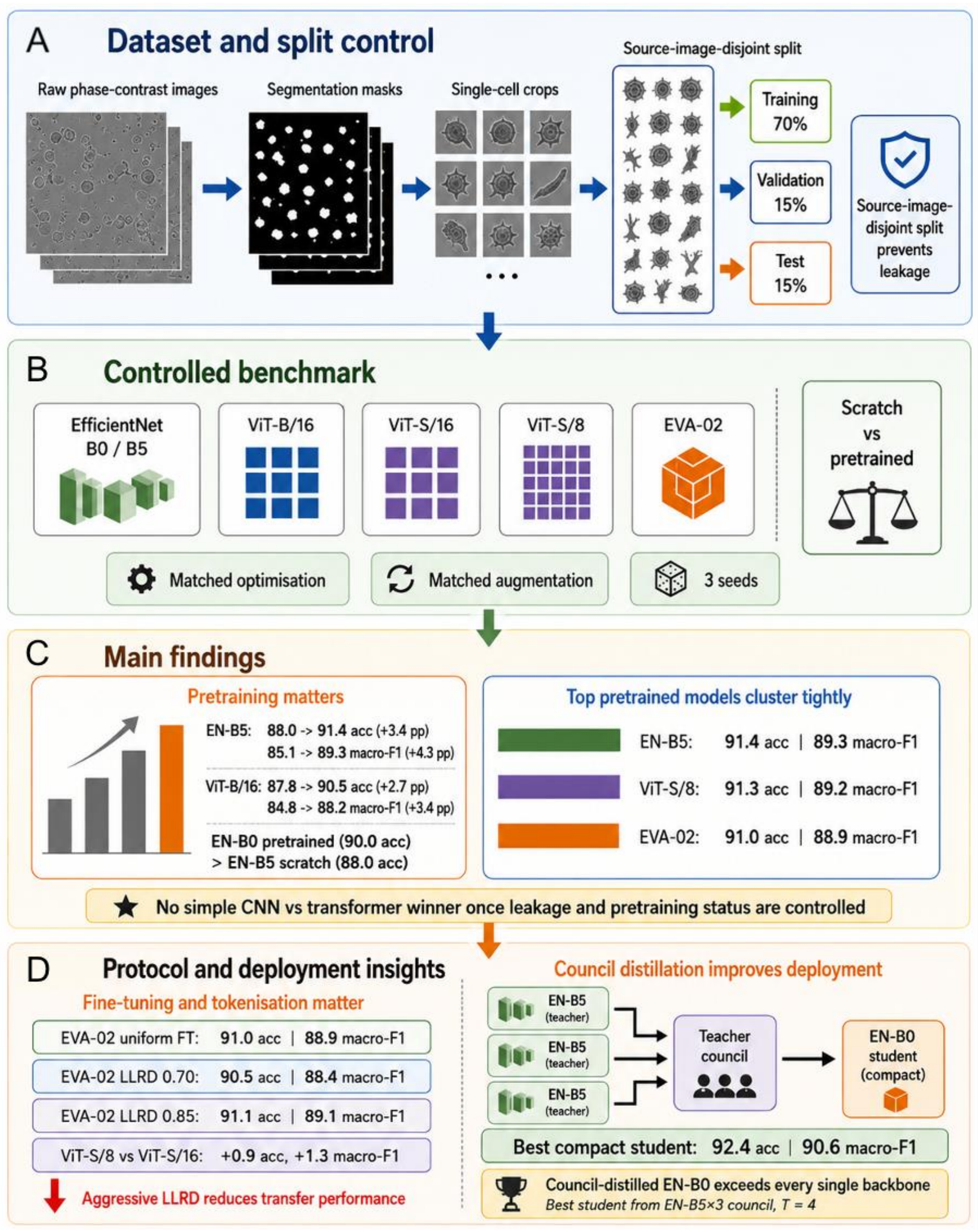

**Graphical Abstract**. A. The benchmarking workflow begins by pairing raw phase-contrast microscopy images with segmentation masks and converting the resulting data into single-cell crops. A source-image-disjoint split is then used to assign cells to training, validation and test sets, ensuring that cells derived from the same source image are not distributed across different partitions and thereby limiting potential data leakage. B. A controlled evaluation framework subsequently compares EfficientNet-B0/B5 (EN-B0/B5), ViT-B/16, ViT-S/16, ViT-S/8 and EVA-02 under matched optimisation and augmentation settings across multiple random seeds, with scratch and pretrained initialisation compared directly for EN-B5 and ViT-B/16. C. Pretraining provides a consistent benefit across architectures, while the strongest pretrained CNN and transformer models show broadly comparable performance, indicating no clear architectural winner under matched transfer-learning conditions. D. Fine-tuning and tokenisation choices further influence the effectiveness of pretrained representations. The deployment strategy uses council knowledge distillation to transfer information from multiple teacher models into a compact EN-B0 student; the strongest student in the present experiments was distilled from the EN-B5×3 council, achieving higher performance than any individually trained backbone while retaining EN-B0 inference cost.

# 1. Introduction

Recent CNN-ViT comparisons in cell imaging reach divergent conclusions, from ViT advantages in explicit transfer-learning studies of white-blood-cell microscopy to CNN or hybrid advantages in broader architectural bake-offs on label-free and microscopic cell datasets (Abou Ali et al., 2023; Fiore et al., 2025; Liu et al., 2025). These discrepancies are difficult to interpret as stable architecture-family effects. ViT performance depends strongly on pretraining and training recipe, and recent microscopy work shows that both self-supervised and regularly pretrained transformer representations can materially change downstream cell-classification performance (Dosovitskiy et al., 2021; Steiner et al., 2021; Moutakanni et al., 2025).

We argue that one important source of apparent disagreement in this literature is a methodological confound: Transformer-family models are not always evaluated under matched or clearly reported pretrained initialisation, even though this is standard practice for competitive ViT transfer (e.g. Fiore et al., 2025; Liu et al., 2025). There remains a broader need for benchmarking of modern architectures on microscopy classification under matched training protocols, with explicit reporting of efficiency metrics and pretraining conditions (Raghu et al., 2021). The present study directly tests whether the reported CNN advantage in label-free single-cell classification survives a controlled benchmark in which pretraining status is matched and made explicit across model families. Using the LIVECell dataset of phase-contrast microscopy images (Edlund et al., 2021) we evaluate representative convolutional and transformer-family models. The choice of LIVECell provides not only a large, standardised benchmark, but also direct comparability with prior architecture evaluations conducted on the same dataset (Fiore et al., 2025).

As the basic unit of life on earth, the cell is of central importance to biology. Single-cell classification is of practical importance as it supports detection of abnormal or undesired phenotypes in medical and biological lab settings. Biologists have a wide array of tools available to them to classify various cell types. However, many of these tests are time intensive and prone to human error. They might also require expensive reagents, labels and specialised microscopes. Label-free classification on the single cell level thus becomes an attractive task for computer vision (Osumi-Sutherland, 2017; Shifat-E-Rabbi et al., 2020; Moutakanni et al., 2025). Automated analysis of microscopy images has become a central enabling technology for modern cellular biology and biomanufacturing (Kraus et al., 2017). In many experimental settings such as high-content screening, perturbation assays, cell line quality control, and longitudinal live-cell studies the primary bottleneck is the reliable conversion of large image volumes into quantitative, reproducible descriptors of cellular identity and state. This is a non-trivial task, as computer models often have to assess minimal differences in cell morphology and texture, confounded by illumination, focus and confluence. Furthermore, datasets are often imbalanced, and domain shift across labs and modalities is severe (Moutakanni et al., 2025). While the task thus appears difficult, computer vision models have been shown to be remarkably effective at it and often doing so while discerning differences invisible even to human experts (e.g. Eckstein et al. 2024).

Neural networks have revolutionised general image recognition and their application to cellular biology is particularly promising for high-throughput analysis. However, it remains unclear which exact neural architectures offer the best accuracy-efficiency trade-off for microscopy-based classification of bioimaging data. Microscopy images differ from natural images in ways that can affect model behaviour, including discriminative cues that are often subtle and local, and labels that can be costly and noisy. Real deployments must furthermore contend with domain shift across instruments, acquisition settings and laboratories (Xing et al., 2017).
Such factors create a plausible advantage for convolutional neural networks (CNNs), whose inductive biases for locality and translation equivariance can yield strong performance in data-limited regimes. Conversely, Vision Transformers (ViTs) and related foundation-model backbones may offer superior representational flexibility and transfer from large-scale pretraining but typically incur higher compute cost and may require more data or careful fine-tuning to realise their benefits (Dosovitskiy, 2021; Steiner et al., 2021; Touvron et al., 2021; Moutakanni et al., 2025).
CNNs provide a scalable approach to object recognition within images by processing local spatial patterns and extracting relevant features (LeCun, Bengio & Hinton, 2015; LeCun et al., 2002). This architecture enables CNNs to learn hierarchical feature representations. The very architecture of typical CNNs with its compression of the feature space into progressively smaller maps encodes inductive biases. In the beginning, context is assessed locally only, in higher layers more complex features are analysed more globally (Battaglia et al., 2018; Dosovitskiy et al., 2021). This broadly aligns with our current understanding of human vision, with basal areas of vision pathways encoding lines or edges, and later neural populations encoding for example specific faces (see e.g. Yamins et al., 2014 and Güçlü et al., 2015).
In contrast to CNNs, ViTs represent an image as a sequence of patches and use self-attention to model interactions across the entire image, which can facilitate learning of global relationships (Dosovitskiy et al., 2021). With less inductive bias, at the level of attention heads, ViTs exhibit a mixture of locality and long-range integration. Early layers contain predominantly local heads alongside a subset of heads that already attend globally, and deeper layers shift toward broadly long-range attention across most heads. This has been quantified by "attention-distance" analyses, analogous to receptive field size in CNNs (Dosovitskiy et al., 2021).
In practice, ViT performance and robustness depend strongly on the training regime. ViTs often benefit substantially from large-scale pretraining and/or data-efficient training strategies (e.g., distillation, augmentation and strong regularisation), and under such settings they can exceed CNN baselines on a range of vision tasks (Dosovitskiy et al., 2021; Touvron et al., 2021; Steiner et al., 2021). For standard training regimes, if no pre-trained data is available, ViTs, compared to CNNs will often underperform with a limited dataset. As transformers they have fewer inductive biases and will need to learn these while training on limited data (Dosovitskiy et al., 2021). Consequently, transfer learning from large-scale pretraining (typically on datasets such as ImageNet) is particularly critical for ViTs and can substantially improve their performance on downstream tasks, often more so than for CNNs (Dosovitskiy et al., 2021; Raghu et al., 2021). This principle has been confirmed specifically for microscopy, where self-supervised pretrained ViTs substantially outperform scratch-trained equivalents (Moutakanni et al., 2025).

The failure to control for pretraining status therefore constitutes a confound that can produce spuriously CNN-favourable conclusions.
We evaluated a number of different architectures: EfficientNet-B0 and EfficientNet-B5 as compact and stronger CNN baselines, ViT-B/16 as a standard Vision Transformer, ViT-S/16 and ViT-S/8 as DINO-pretrained transformers differing only in patch size, and EVA-02 as a modern pretrained ViT-family backbone. We further tested whether an EVA-02 fine-tuning recipe based on layer-wise learning-rate decay transfers to phase-contrast microscopy, and whether offline knowledge distillation can transfer complementary information from larger or diverse teachers into a compact EfficientNet-B0 student. Distillation offers a practical route to compact, high-accuracy models necessary for high-throughput application in biological microscopy labs.
The study has four aims. First, we test whether a reported CNN advantage on LIVECell persists under source-image-disjoint, matched-pretraining evaluation. Second, we quantify the extent to which pretraining, rather than architecture family alone, explains performance differences. Third, we assess whether transformer tokenisation and fine-tuning strategy materially affect phase-contrast transfer. Fourth, we evaluate whether teacher and council distillation can decouple inference efficiency from the architecture of the highest-performing backbone.
We hypothesise that the CNN versus ViT gap will narrow once pretraining and split protocol are controlled, that fine-tuning recipes developed for natural-image transfer will not necessarily transfer unchanged to phase-contrast microscopy, and that compact students can recover or exceed large-backbone performance when trained from complementary teacher signals.

# 2. Results

### 2.1 Main Benchmark: Pretrained Models

All experiments used single-cell crops extracted from LIVECell, a large public phase-contrast microscopy dataset (Edlund et al., 2021). To eliminate parent-image leakage, crops are partitioned by source image into disjoint train, validation, and test splits, and optimisation, augmentation, and evaluation protocols are held constant across all backbones. We compare six representative architectures spanning both families: EfficientNet-B0 and EfficientNet-B5 as compact and stronger convolutional baselines, ViT-B/16 as a standard Vision Transformer, ViT-S/16 and ViT-S/8 as DINO-pretrained transformers differing only in patch size, and EVA-02 as a modern pretrained ViT-family backbone. Unless stated otherwise, all models use pretrained initialisation with uniform end-to-end fine-tuning, and all results are reported as mean ± SD across three random seeds. Full architecture specifications, training schedules, optimiser settings, augmentation, and the distillation and profiling protocols are given in the Section 4 (Methods).
Under these conditions, the strongest backbones from both architecture families occupied a narrow performance range rather than separating into a clear convolutional-versus-transformer hierarchy. Full classification results are reported in Table 1 and summarised in Figure 1. EfficientNet-B5 was nominally highest, reaching 91.35 ± 0.18% accuracy and 89.3 ± 0.2% macro-F1. ViT-S/8 followed closely at 91.27 ± 0.16% accuracy and 89.18 ± 0.18% macro-F1, while EVA-02 reached 91.03 ± 0.14% accuracy and 88.9 ± 0.2% macro-F1.

ViT-B/16 reached 90.5 ± 0.5% accuracy and 88.2 ± 0.6% macro-F1, while ViT-S/16 reached 90.4 ± 0.4% accuracy and 87.9 ± 0.6% macro-F1. The compact EfficientNet-B0 baseline reached 90.00 ± 0.19% accuracy and 87.7 ± 0.2% macro-F1.

No statistically significant difference was detected between EfficientNet-B5 and either of the leading transformer-family models in two-sided paired t-tests over the three seeds. EfficientNet-B5 versus ViT-S/8 gave $p = 0.069$ for accuracy and $p = 0.076$ for macro-F1, while EfficientNet-B5 versus EVA-02 gave $p = 0.202$ for accuracy and $p = 0.173$ for macro-F1. Given the small number of seeds, these tests should be interpreted as descriptive evidence that no stable family-level separation was observed rather than as proof of performance equivalence.

Tokenisation granularity emerged as a more specific architectural factor. Under matched model width and DINO pretraining, ViT-S/8 outperformed ViT-S/16 by 0.9 percentage points in accuracy and 1.3 points in macro-F1. This comparison differs principally in patch size and therefore provides the cleanest evidence that finer tokenisation benefits small single-cell crops. ViT-S/8 also outperformed the larger ViT-B/16 by 0.8 percentage points in accuracy and 1.0 point in macro-F1 while using approximately one quarter of its parameters. The two models nevertheless had comparable computational cost, because the longer patch-8 token sequence offset the smaller model width. Finer tokenisation therefore improved predictive performance but did not reduce computation.

**Table 1. Classification performance of pretrained architectures.** Accuracy, balanced accuracy, macro-F1, and expected calibration error are reported as mean ± SD across three random seeds. Models are ordered by balanced accuracy.

| Model | Family | Accuracy (%) | Balanced accuracy (%) | Macro-F1 (%) | ECE (%) |
|---|---|---|---|---|---|
| **EN-B5** | CNN | 91.35 ± 0.18 | 88.96 ± 0.15 | 89.3 ± 0.2 | 2.6 ± 0.2 |
| **EVA-02** | ViT | 91.03 ± 0.14 | 88.9 ± 0.3 | 88.9 ± 0.2 | 2.87 ± 0.10 |
| **ViT-S/8 DINO** | ViT | 91.27 ± 0.16 | 88.6 ± 0.3 | 89.18 ± 0.18 | 2.0 ± 0.2 |
| **ViT-B/16** | ViT | 90.5 ± 0.5 | 87.9 ± 0.5 | 88.2 ± 0.6 | 2.20 ± 0.13 |
| **ViT-S/16 DINO** | ViT | 90.4 ± 0.4 | 87.6 ± 0.8 | 87.9 ± 0.6 | 2.5 ± 0.6 |
| **EN-B0** | CNN | 90.00 ± 0.19 | 87.5 ± 0.3 | 87.7 ± 0.2 | 2.3 ± 0.3 |

**Figure 1. Classification performance and effect of pretrained initialisation.** (A) Accuracy, balanced accuracy and macro-F1 for the pretrained architectures. (B) Balanced accuracy and macro-F1 for EfficientNet-B5 and ViT-B/16 under random and pretrained initialisation. Bars show means across three random seeds and error bars indicate one standard deviation. Values above pretrained bars denote improvements over random initialisation. CNN models are shown in red and transformer-family models in gold.

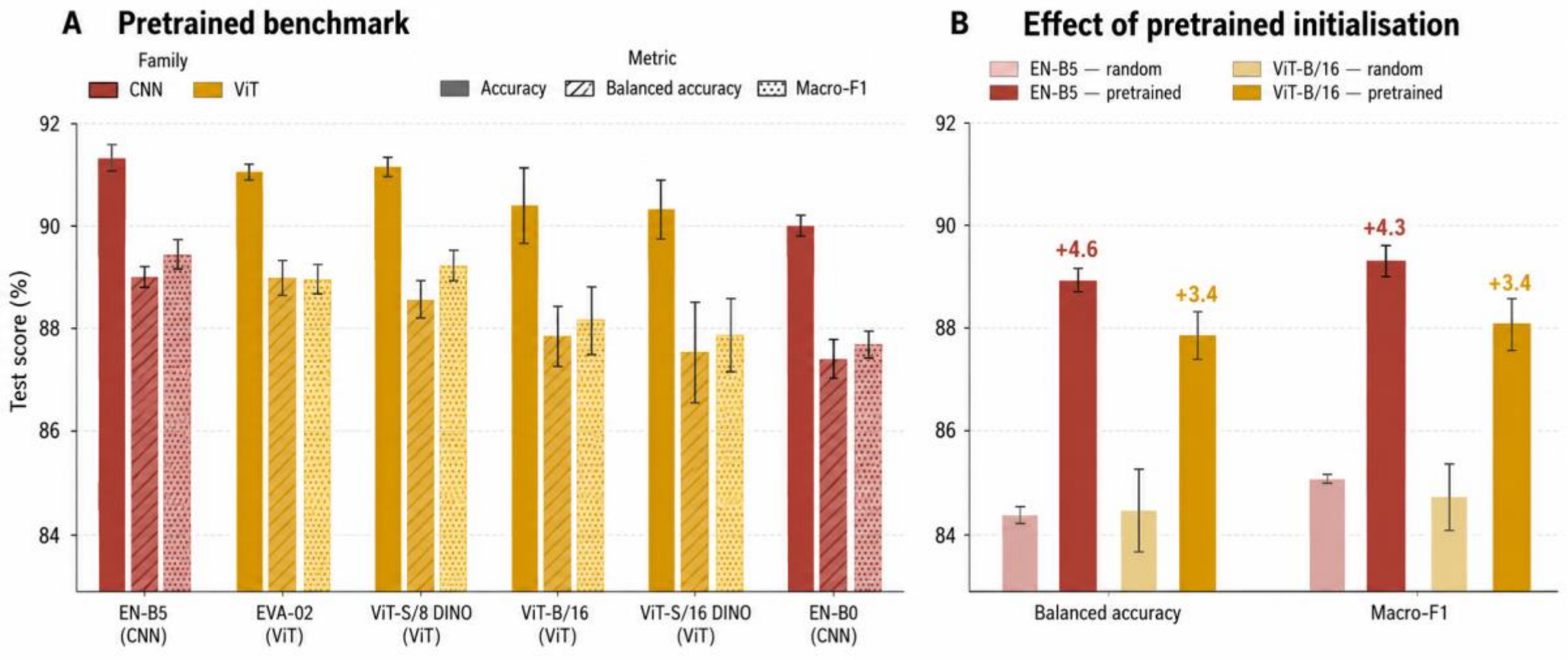

## 2.2 Pretraining versus random initialisation

To isolate the effect of pretraining from architecture family, EfficientNet-B5 and ViT-B/16 were evaluated under both pretrained and random initialisation. Data splits, augmentation, optimiser, effective batch size, and evaluation procedures were held constant. Randomly initialised models were given a longer training schedule, as described in Section 4 (Methods).

Pretraining produced substantial improvements in both architecture families. EfficientNet-B5 improved from 87.99 ± 0.10% accuracy and 85.07 ± 0.03% macro-F1 under random initialisation to 91.35 ± 0.18% accuracy and 89.3 ± 0.2% macro-F1 under pretrained initialisation. This corresponded to gains of 3.4 percentage points in accuracy and 4.3 points in macro-F1.

ViT-B/16 showed the same pattern. Randomly initialised ViT-B/16 reached 87.8 ± 0.3% accuracy and 84.8 ± 0.4% macro-F1, whereas pretrained ViT-B/16 reached 90.5 ± 0.5% accuracy and 88.2 ± 0.6% macro-F1. This corresponded to gains of 2.7 percentage points in accuracy and 3.4 points in macro-F1.

The corresponding changes in accuracy and macro-F1 are reported in Table 2. Figure 1B shows changes in balanced accuracy and macro-F1. The magnitude of the pretraining effect was sufficient to alter model-scale rankings. The compact pretrained EfficientNet-B0 reached 90.00% accuracy and therefore outperformed the substantially larger scratch-trained EfficientNet-B5, which reached 87.99%. Initialisation regime can consequently dominate model scale in this setting.

Non-pretrained models also showed little intrinsic separation between the two architecture families. Randomly initialised EfficientNet-B5 and ViT-B/16 reached 87.99% and 87.8% accuracy, respectively. The difference between them was much smaller than the improvement produced by pretraining either architecture.

**Table 2. Random initialisation versus pretrained comparison.** Values are mean ± SD across three random seeds. Δ = pretrained − random; positive values indicate a pretraining gain.

| Model | Random acc. (%) | Pretrained acc. (%) | Δ acc. | Random macro-F1 | Pretrained macro-F1 | Δ macro-F1 |
|---|---|---|---|---|---|---|
| **EN-B5** | 87.99 ± 0.10 | 91.35 ± 0.18 | +3.4 pp | 85.07 ± 0.03 | 89.3 ± 0.2 | +4.3 pp |
| **ViT-B/16** | 87.8 ± 0.3 | 90.5 ± 0.5 | +2.7 pp | 84.8 ± 0.4 | 88.2 ± 0.6 | +3.4 pp |

### 2.3 Layer-wise learning-rate decay

The main pretrained benchmark used uniform fine-tuning. To test whether a fine-tuning strategy developed for natural-image transfer carries over unchanged to phase-contrast microscopy, EVA-02 was additionally fine-tuned using layer-wise learning-rate decay. We evaluated a decay factor of 0.70, corresponding to the original EVA-02 fine-tuning recipe, and a milder factor of 0.85. This was treated as a recipe-transfer experiment rather than as a broad hyperparameter search. For more details see Section 4 (Methods).

Uniform fine-tuning, corresponding to an LLRD factor of 1.0, reached 91.03 ± 0.14% accuracy, 88.9 ± 0.2% macro-F1, and 88.9 ± 0.3% balanced accuracy. LLRD with a factor of 0.70 reached 90.51 ± 0.16% accuracy, 88.4 ± 0.3% macro-F1, and 88.2 ± 0.2% balanced accuracy. This represented reductions of 0.5 percentage points in accuracy, 0.5 points in macro-F1, and 0.7 points in balanced accuracy.

The reduction was observed across all three seeds. Under two-sided paired t-tests, the reduction in accuracy was statistically significant, $t(2) = 5.44$, $p = 0.032$. Macro-F1 and balanced accuracy showed reductions in the same direction but did not reach conventional significance: macro-F1, $t(2) = 2.64$, $p = 0.118$; balanced accuracy, $t(2) = 3.45$, $p = 0.075$.

The effect depended on the strength of the decay. EVA-02 with the milder factor of 0.85 reached 91.1 ± 0.2% accuracy, 89.00 ± 0.21 balanced accuracy and 89.1 ± 0.2% macro-F1, which was similar to uniform fine-tuning. The result therefore does not show that LLRD is generally harmful, but the decay value transferred from the original natural-image recipe reduced performance in this phase-contrast setting, whereas milder decay was approximately neutral.

### 2.4 Knowledge Distillation

Finally, we evaluated whether compact student models could inherit useful structure from stronger or complementary teachers. We distilled compact EfficientNet-B0 students from stronger teachers, training on a weighted combination of ground-truth labels and temperature-scaled teacher predictions. We evaluated single-teacher distillation (from either EfficientNet-B5 or EVA-02) and council distillation, in which the logits of several teachers are averaged sample-wise before temperature scaling. Three councils were compared: homogeneous EVA-02×3, mixed 2×EVA-02 + EfficientNet-B5, and homogeneous EfficientNet-B5×3. Full loss formulation, temperatures, and weighting are given in Section 4 (Methods). Full results for all teachers, council compositions and temperatures are reported in Table 3.

The pretrained EfficientNet-B0 baseline reached 90.00 ± 0.19% accuracy and 87.7 ± 0.2% macro-F1. Single-teacher distillation substantially improved this compact model. Distillation from EfficientNet-B5 reached 91.44 ± 0.13% accuracy and 89.46 ± 0.11% macro-F1 at T=3, and 91.5 ± 0.2% accuracy and 89.6 ± 0.3% macro-F1 at T=4. The latter represented improvements of 1.5 percentage points in accuracy and 1.9 points in macro-F1 over the undistilled baseline.

EVA-02 was also an effective teacher. EVA-02-to-EfficientNet-B0 distillation reached 91.29 ± 0.12% accuracy and 89.23 ± 0.16% macro-F1 at T=3, and 91.42 ± 0.19% accuracy and 89.4 ± 0.3% macro-F1 at T=4. Transformer-family teachers could therefore transfer useful predictive structure into a strong but compact convolutional student, even though EVA-02 was not the strongest standalone backbone.

Council distillation produced the largest improvements. The homogeneous EVA-02×3 council reached approximately 92.16 ± 0.03% accuracy and 90.30 ± 0.05% macro-F1 at T=3 and 92.14 ± 0.06% accuracy with 90.28 ± 0.11% macro F1 at T=4. The mixed 2×EVA-02 + EfficientNet-B5 council reached 92.28 ± 0.04% accuracy and 90.44 ± 0.05% macro-F1 at T=3, and 92.36 ± 0.01% accuracy and 90.54 ± 0.03% macro-F1 at T=4.

The homogeneous EfficientNet-B5×3 council was numerically strongest. It reached 92.30 ± 0.04% accuracy and 90.51 ± 0.03% macro-F1 at T=3, and 92.40 ± 0.06% accuracy and 90.62 ± 0.09% macro-F1 at T=4. Relative to the pretrained EfficientNet-B0 baseline, the best council-distilled student improved accuracy by 2.4 percentage points and macro-F1 by 2.9 points. It also exceeded the best single-teacher-distilled student by 0.9 percentage points in accuracy and 1.1 points in macro-F1.

The council-distilled EfficientNet-B0 student surpassed every individually trained backbone in the main benchmark while retaining the EfficientNet-B0 inference architecture. Homogeneous EVA-02, mixed EVA-02/EfficientNet-B5, and homogeneous EfficientNet-B5 councils all substantially outperformed single-teacher distillation, while the homogeneous EfficientNet-B5 council was numerically strongest.

Direct council inference established the performance attainable without compression into a student, as shown in Table 4. Logit averaging across the three EVA-02 teachers reached 92.55% accuracy and 90.76% macro-F1, while the mixed council reached 92.71% and 90.98%, respectively. The EfficientNet-B5 council reached 92.70% accuracy and the highest direct-council macro-F1 of 91.00%. Direct council inferences however have additive inference costs, as shown in Table 5.

Interestingly, the fraction of the direct-council performance gain retained by the EfficientNet-B0 student increased numerically from the homogeneous EVA-02 council to the mixed council and the homogeneous EfficientNet-B5 council. This ordering is consistent with the possibility that teacher–student architectural alignment facilitates compression of the teacher decision function.

**Table 3. Knowledge distillation results.** Baseline row shows the pretrained EfficientNet-B0 student fine-tuned without distillation. All distillation rows use the same pretrained EfficientNet-B0 student architecture. Δ values report gains over the baseline student. Council distillation averages multiple teacher logits before temperature scaling. Values are mean ± SD across three random seeds.

| Setting | Teacher signal | Temperature | Accuracy (%) | **Δ Acc.** | Macro-F1 | **Δ F1** |
|---|---|---|---|---|---|---|
| **Baseline** | None | n/a | 90.00 ± 0.19 | n/a | 87.7 ± 0.2 | n/a |
| **Single teacher** | EN-B5 | 3 | 91.44 ± 0.13 | 1.44 | 89.46 ± 0.11 | 1.75 |
| **Single teacher** | EN-B5 | 4 | 91.5 ± 0.2 | 1.54 | 89.6 ± 0.3 | 1.86 |
| **Single teacher** | EVA-02 | 3 | 91.29 ± 0.12 | 1.29 | 89.23 ± 0.16 | 1.52 |
| **Single teacher** | EVA-02 | 4 | 91.42 ± 0.19 | 1.42 | 89.4 ± 0.3 | 1.69 |
| **Council** | EVA-02x3 | 3 | 92.16 ± 0.03 | 2.16 | 90.30 ± 0.05 | 2.59 |
| **Council** | EVA-02x3 | 4 | 92.14 ± 0.06 | 2.14 | 90.28 ± 0.11 | 2.57 |
| **Council** | 2xEVA-02 + EN-B5 | 3 | 92.28 ± 0.04 | 2.28 | 90.44 ± 0.05 | 2.73 |
| **Council** | 2xEVA-02 + EN-B5 | 4 | 92.36 ± 0.01 | 2.36 | 90.54 ± 0.03 | 2.83 |
| **Council** | EN-B5x3 | 3 | 92.30 ± 0.04 | 2.3 | 90.51 ± 0.03 | 2.8 |
| **Council** | EN-B5x3 | 4 | 92.40 ± 0.06 | 2.4 | 90.62 ± 0.09 | 2.91 |

**Table 4. Council baseline and council distillation comparison.** Direct councils display higher accuracies, which only partially translate into student performance, the distillation gap differs across council structure. Council distillation averages multiple teacher logits before temperature scaling.

| Teacher council | Direct accuracy | Direct macro-F1 | Best distilled accuracy | Best distilled macro-F1 | Distillation gap, acc./F1 |
|---|---:|---:|---:|---:|---:|
| **EVA-02×3** | 92.55 | 90.76 | 92.16 ± 0.03, T=3 | 90.30 ± 0.05 | 0.39 / 0.46 pp |
| **2×EVA-02 + EN-B5** | **92.71** | 90.98 | 92.36 ± 0.01, T=4 | 90.54 ± 0.03 | 0.35 / 0.44 pp |
| **EN-B5×3** | 92.70 | **91.00** | **92.40 ± 0.06**, T=4 | **90.62 ± 0.09** | 0.30 / 0.38 pp |

**Table 5. Council versus EN-B0 size comparison.** Rounded comparison of values given compared to EN-B0 as baseline.

| Council | Aggregate parameters | Aggregate GMACs | Versus EN-B0 GMACs |
|---|---|---|---|
| EVA-02×3 | 257.28 M | 65.82 | 173× |
| 2×EVA-02 + EN-B5 | 199.88 M | 46.23 | 122× |
| EN-B5×3 | 85.08 M | 7.05 | 19× |
| Distilled EN-B0 | 4.02 M | 0.38 | 1× |

## 2.5 Computational complexity and inference efficiency

Model complexity and A100 inference measurements are reported in Table 6. Model complexity is reported as parameter count and GMACs (forward FLOPs halved to multiply-adds); inference is profiled on a single NVIDIA A100 GPU at 224 × 224 with bfloat16 autocast. Distilled students are evaluated using student inference cost only. Full profiling configuration is given in Section 4 (Methods). Parameter counts ranged from 4.0 million for EfficientNet-B0 to 85.8 million for ViT-B/16, while computation ranged from 0.38 GMACs for EfficientNet-B0 to 21.94 GMACs for EVA-02.

At batch size 64, ViT-S/16 achieved the highest throughput, processing approximately 7380 images/s with 317.7 MB peak inference memory. EfficientNet-B0 was the smallest and lowest-GMAC model and processed approximately 5960 images/s with 430.9 MB peak memory. ViT-B/16 processed approximately 3,490 images/s, while ViT-S/8 processed approximately 2150 images/s. EfficientNet-B5 and EVA-02 were slower in this profiling configuration, processing approximately 1670 and 1320 images/s, respectively.

The ViT-S/8 versus ViT-S/16 comparison demonstrated the computational cost of finer tokenisation. Although both models contained 21.7 million parameters, reducing the patch size from 16 × 16 to 8 × 8 increased computation from 4.24 to 16.73 GMACs and peak inference memory from 317.7 to 760.7 MB. Batch-64 throughput decreased from approximately 7380 to 2150 images/s. The predictive advantage of finer tokenisation was therefore accompanied by a substantial compute, memory, and throughput penalty.

These measurements also show that parameter count and GMACs do not directly determine realised GPU speed. ViT-S/16 achieved greater A100 throughput than EfficientNet-B0 despite requiring substantially more nominal computation, while EfficientNet-B5 had relatively low GMACs but comparatively poor throughput. As shown in Figure 2, distillation allows very high performance at low cost. The council-distilled student retains EfficientNet-B0 throughput (≈5960 img/s) while exceeding the macro-F1 of every individually trained model, lifting the attainable frontier by 2.9 points at unchanged inference cost.

**Table 6. Model complexity and A100 inference profiling.** Parameter counts and GMACs are reported for 224×224 inputs. Inference profiling used synthetic 224×224 tensors, bfloat16 autocast, torch.inference_mode(), CUDA event timing, cuDNN benchmark mode, and a single A100SXM GPU. Models are ordered by batch-64 throughput.

| Model | Family | Params (M) | GMACs | A100 bs1 latency (ms/img) | A100 bs64 throughput (img/s) | A100 bs64 latency (ms/img) | A100 peak memory (MB) |
|---|---|---|---|---|---|---|---|
| **ViT-S/16 DINO** | ViT | 21.67 | 4.24 | 4.26 | 7380 | 0.136 | 317.7 |
| **EN-B0** | CNN | 4.02 | 0.38 | 7.82 | 5960 | 0.168 | 430.9 |
| **ViT-B/16** | ViT | 85.80 | 16.85 | 4.27 | 3490 | 0.287 | 840.1 |
| **ViT-S/8 DINO** | ViT | 21.67 | 16.73 | 4.42 | 2150 | 0.466 | 760.7 |
| **EN-B5** | CNN | 28.36 | 2.35 | 18.35 | 1670 | 0.598 | 707.0 |
| **EVA-02** | ViT | 85.76 | 21.94 | 9.37 | 1320 | 0.759 | 1075.9 |

**Figure 2. Performance–throughput trade-off among pretrained backbones and the council-distilled student.** Mean macro-F1 against batch-64 throughput on a single NVIDIA A100SXM GPU (224 × 224 synthetic inputs, bfloat16 autocast). Bubble area denotes parameter count; colour denotes architecture family (CNN, red; transformer, gold). The dashed line is the Pareto frontier over individually trained backbones, computed from means without uncertainty. EfficientNet-B0 student distilled from the EfficientNet-B5×3 council (T = 4) is marked by a diamond with an arrow showing its gain over the undistilled EfficientNet-B0 at identical inference cost. Error bars show one SD across three seeds.

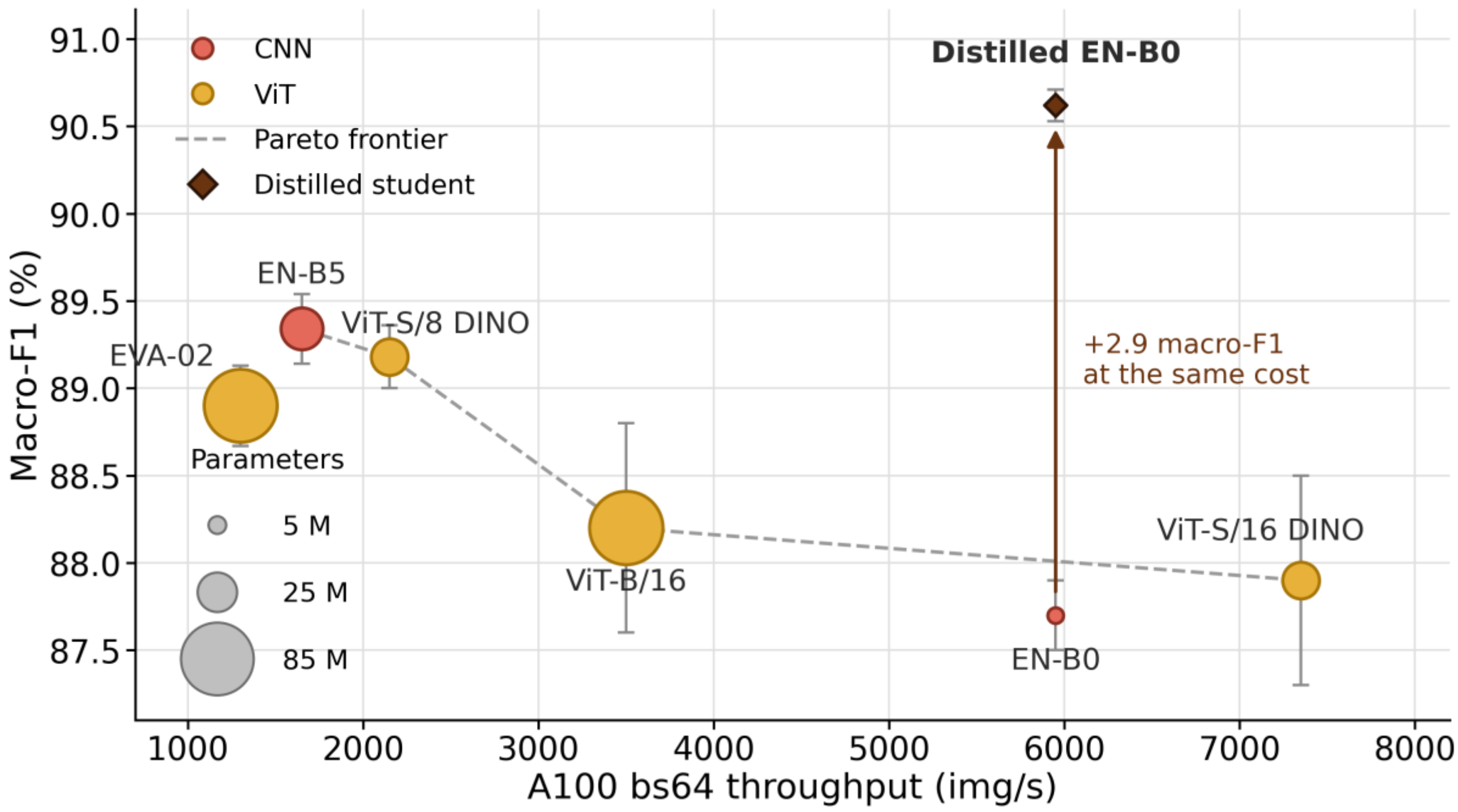

# 3. Discussion

### 3.1 Pretraining and architecture-family comparisons

The central finding of this study is that the previously reported large CNN advantage for LIVECell single-cell classification does not survive controlled experimental conditions. Fiore et al. interpreted this advantage as evidence that local convolutional inductive biases are better matched to the discriminative cues of phase-contrast microscopy than the global attention mechanisms of ViTs. The corrected results show that this interpretation is incomplete, because it does not isolate architecture family from initialisation regime.

Pretraining produced large gains in both families. EfficientNet-B5 improved from 88% accuracy (85.1 macro-F1) when trained from random initialisation to 91.4% (89.3) with pretrained weights, a gain of 3.4 percentage points; ViT-B/16 showed the same qualitative pattern, improving from 87.8% (84.8) to 90.5% (88.2), a gain of 2.7 points. These within-architecture comparisons show that pretraining is a first-order determinant of downstream performance. The effect is large enough to override model scale: pretrained EfficientNet-B0 outperformed scratch-trained EfficientNet-B5 despite being the smaller and weaker CNN. Initialisation can therefore dominate architecture scale, and comparisons that do not match or clearly report pretraining status cannot be interpreted as pure architecture-family comparisons. This regime dependence is what narrows the apparent gap between the families, as shown in Table 1. On theoretical grounds, convolutional models should have an advantage in lower-data or scratch-training regimes, because locality and translation equivariance are built into the architecture and therefore do not need to be learned entirely from data. The results are consistent with this expectation, but only weakly: when trained from random initialisation, EfficientNet-B5 and ViT-B/16 performed almost identically (88% versus 87.8%). Convolutional inductive bias remains useful, but the scratch results do not support a strong family-level claim that phase-contrast microscopy intrinsically favours CNNs over transformer-based models. Under random initialisation, EfficientNet-B5 and ViT-B/16 produced similar performance (Table 2), and under pretrained initialisation, no statistically significant separation was detected between EfficientNet-B5 and the leading transformer-family models (Table 1). The results therefore do not support the family-level claim that phase-contrast microscopy intrinsically favours CNNs. LIVECell is large by microscopy standards, but not large enough to train high-capacity ViTs from random initialisation to full representational capacity; dataset size is relative to model class, and ImageNet pretraining supplies useful transferable structure to both CNNs and ViTs without producing a stable transformer advantage.

These results suggest that inductive bias should not be treated as a binary family property. Modern vision architectures differ not only in whether they use convolution or self-attention, but also in tokenisation strategy, parameterisation, optimisation behaviour, pretraining objective, fine-tuning recipe, and effective representational capacity. ViT-S/8 is particularly informative here: despite being smaller than ViT-B/16, it outperformed it by 0.8 percentage points in accuracy and 1 point in macro-F1, suggesting that for small single-cell crops, preserving fine-grained local morphology through finer patch tokenisation may matter more than increasing model scale.

The corrected benchmark therefore changes the interpretation of the earlier CNN-ViT gap without dismissing convolutional bias. EfficientNet-B5 remained highly competitive and was still the best single pretrained backbone; the point is that the contribution of inductive bias cannot be interpreted independently of training regime. Pretraining status is not a minor implementation detail but part of the experimental object being compared. A model evaluated in the transfer-learning regime in which it is normally used is not the same experimental object as the same architecture trained from random initialisation. The corrected conclusion is that architecture-family rankings are fragile unless pretraining, split protocol, tokenisation, and fine-tuning recipe are controlled and reported explicitly.

This is unlikely to be a LIVECell-specific issue. Moutakanni et al. (2025) independently showed that pretrained or self-supervised transformer representations substantially improve downstream microscopy classification relative to scratch-trained equivalents, and the convergence across microscopy settings supports the broader inference that pretraining regime is a primary experimental variable in biological-image architecture benchmarks.

We therefore recommend that future benchmarks report pretraining status, pretraining domain, split granularity, fine-tuning protocol, and tokenisation alongside parameter count, GMACs, accuracy, calibration, memory, throughput, and per-class reliability. Finally, this critique should be separated from the other contributions of Fiore et al.: their Tensor Network module remains a distinct contribution to interpretable cell classification, and their ablation against parameter-matched MLPs addresses a separate question from the architecture-family comparison. Those contributions do not depend on the CNN-ViT ranking claim, and a natural follow-up would be to re-evaluate such modules using properly pretrained ViT and CNN backbones under source-image-disjoint splits.

### 3.2 Domain shift and tokenisation

EfficientNet-B5 had the numerically highest mean performance among the individual backbones, but no statistically significant separation was detected from the leading transformer-family models (Table 1). EfficientNet-B5 reached $91.35 \pm 0.18\%$ accuracy and $89.3 \pm 0.2\%$ macro-F1, compared with $91.27 \pm 0.16\%$ and $89.18 \pm 0.18\%$, respectively, for ViT-S/8. The paired comparisons did not reach conventional statistical significance ($p = 0.069$ for accuracy and $p = 0.076$ for macro-F1). These results therefore provide no evidence of residual CNN superiority under pretrained initialisation.

The first factor is domain shift. ImageNet pretraining supplies representations learned from natural images: colour, object-level structure, texture, background context, and naturalistic spatial statistics. Phase-contrast microscopy is near-greyscale, dominated by membrane contrast, diffraction halos, focus variation, confluence, and subtle intracellular texture. Even lower-level source-domain representations may therefore map imperfectly to the target domain. When comparing between architecture families, CNNs may be less dependent on pretraining for spatial structure because locality and translation equivariance are already encoded architecturally. ViTs, by contrast, rely more heavily on pretraining to supply spatial and representational scaffolding. If that scaffolding is domain-mismatched, transfer may be less efficient.

The layer-wise learning-rate decay (LLRD) result provides targeted support for this interpretation. LLRD is often beneficial for ViT fine-tuning because it protects lower layers assumed to encode generic transferable features while allowing higher layers to adapt more strongly. In the present setting, the original EVA-02 LLRD factor of 0.7 reduced accuracy from 91% to 90.5% and macro-F1 from 88.9 to 88.4. Balanced accuracy also decreased. This suggests that the early layers protected by aggressive LLRD may be precisely the layers that require adaptation under phase-contrast domain shift. Here the original EVA-02 natural-image fine-tuning recipe transfers poorly to phase-contrast microscopy. The pattern is consistent with aggressive decay constraining adaptation to the modality's distinct intensity, texture, edge, and halo statistics.

The second factor of within ViT family differences is tokenisation. At 224 × 224 resolution, ViT-B/16 represents the image using a 14 × 14 grid of 196 tokens. For small single-cell crops, discriminative information may be concentrated in membrane texture, nuclear granularity, cytoplasmic organisation, boundary irregularity, and local halo structure; coarse tokenisation may compress these cues before self-attention can integrate them. The strongest evidence comes from the matched ViT-S comparison: under the same model scale and DINO pretraining, ViT-S/8 outperformed ViT-S/16 by 0.9 percentage points in accuracy and 1.3 points in macro-F1. ViT-S/8 also exceeded ViT-B/16 by 0.8 points in accuracy and 1.0 point in macro-F1 despite using approximately one quarter as many parameters. Although the latter is not an isolated patch-size comparison, together these results identify tokenisation granularity as a consequential design variable for small, texture-rich biological objects.

Under matched model width and DINO pretraining, finer tokenisation produced higher mean performance for ViT-S/8 than for ViT-S/16 and brought the smaller transformer into the same performance range as EfficientNet-B5 (Table 1). Because no statistically significant separation was detected between ViT-S/8 and EfficientNet-B5, their remaining numerical ordering should not be interpreted mechanistically.

A possible interpretation of these observations is that the advantage of global contextual modelling may be limited for the cell crops considered here. Unlike many natural-image classification problems, a LIVECell crop is largely object-centric: the cell typically occupies much of the visual field, and the surrounding context is relatively limited. Although global structure is not entirely absent, the crop provides less scene-level information of the kind that can be exploited through long-range interactions in natural images. Attention-distance analyses of ImageNet-pretrained ViTs show that early layers contain both local and relatively global attention heads (Dosovitskiy et al., 2021). In natural images, such global interactions can potentially exploit relationships between an object and its surrounding scene. For the cell crops used in this study, however, the most informative cues may instead be predominantly local or object-internal, such as membrane texture, halo profile and nuclear granularity, together with whole-cell morphology.

From this perspective, the similar performance of EfficientNet-B5 and ViT-S/8 could reflect the fact that both architectures are capable of representing information at spatial scales appropriate for the task, albeit through different mechanisms. Convolutional networks impose locality through their architectural structure, whereas finer patch tokenisation allows a ViT to represent similarly local information at the token level. Under phase-contrast domain shift, a pretrained ViT may nevertheless need to adapt these representations to the modality-specific intensity and edge statistics. The EVA-02 LLRD result is consistent with the possibility that restricting adaptation in early layers is detrimental when these low-level representations require substantial adjustment. Conversely, if the discriminative information in these crops is primarily local and object-internal, global attention may provide less additional benefit than it does in scene-based natural-image classification. On this interpretation, the relevant distinction may therefore be less about attention versus convolution per se, and more about the spatial scale at which each architecture represents information and how readily those representations can adapt to the target domain.

This remains an interpretation rather than a demonstrated mechanism. In particular, the patch-size comparison does not isolate locality from sequence length and computational cost, and the upsampling of crops to 224 × 224 means that patch size determines granularity relative to cellular structure rather than directly determining pixel-level resolution. A direct test would examine attention-distance distributions in the fine-tuned models. One hypothesis is that early-layer attention would become more local after fine-tuning on cell crops, potentially with this effect depending on the strength of layer-wise learning-rate decay. More generally, comparing attention-distance distributions, attention entropy and feature similarity across pretrained, uniformly fine-tuned and strongly layer-decayed models could establish whether the observed performance differences are accompanied by measurable changes in spatial representation.

### 3.3 Knowledge distillation

The distillation results (see Table 3) extend the study beyond architecture comparison and show that the best deployment model need not be the best standalone backbone. EfficientNet-B0, the compact student, reached 90% accuracy and 87.7 macro-F1 under standard pretrained fine-tuning. Distillation substantially improved this baseline. With EfficientNet-B5 as teacher, the student reached 91.4% accuracy and 89.5 macro-F1 at T = 3, and 91.5% accuracy and 89.6 macro-F1 at T = 4. EVA-02 was also an effective teacher, reaching 91.3% accuracy and 89.2 macro-F1 at T = 3, and 91.4% accuracy and 89.4 macro-F1 at T = 4.

EfficientNet-B5 is both the stronger standalone backbone and the slightly stronger single teacher. Both teacher families transfer useful information to a compact CNN student, showing that teacher accuracy alone remains an incomplete proxy for teacher utility.

The strongest mean results were obtained through council distillation (Table 3). All council configurations achieved higher mean accuracy and macro-F1 than any individual backbone while retaining the inference cost of the compact EfficientNet-B0 student. The numerically highest values were obtained with the homogeneous EfficientNet-B5 council at T = 4, which reached 92.40 ± 0.06% accuracy and 90.62 ± 0.09% macro-F1. However, these values were only 0.04 and 0.08 percentage points higher, respectively, than those of the mixed council at T = 4; this small numerical separation does not substantiate superiority of one council composition. The principal result is therefore the advantage associated with council distillation, rather than a reliable effect of teacher family.

The most parsimonious interpretation is that council distillation transfers information that is not captured by hard labels or by a single teacher. Temperature-scaled soft targets encode class-similarity structure, uncertainty, and non-target probability mass, what Hinton et al. termed dark knowledge. Averaging teacher logits can also reduce idiosyncratic errors and smooth over model-specific overconfidence. The strong performance of both homogeneous and mixed councils indicates that the gain is primarily associated with aggregating multiple teacher signals, whereas architectural diversity is not required for council distillation to be effective.

Direct councils outperformed their corresponding council-distilled students by 0.30–0.39 percentage points in accuracy and 0.38–0.46 points in macro-F1. Nevertheless, the distilled EfficientNet-B0 students retained approximately 85–89% of the direct councils' improvement over the undistilled EfficientNet-B0 baseline while requiring only a single EfficientNet-B0 forward pass. Direct ensembling therefore maximised predictive performance, whereas council distillation recovered most of this benefit at substantially lower inference cost. The largest ensemble, three EVA-02 reached 173 times higher GMACs than EN-B0, while being 64 times larger in parameters, as seen in Table 5.
The practical implication is that the deployment frontier is not determined solely by choosing the strongest standalone backbone. A compact student can exceed larger individual models when trained from council teacher signals. This partially decouples representational quality from inference architecture: expensive models can be used during training to shape a cheap model used at deployment. For high-throughput microscopy, where inference cost and integration into laboratory pipelines matter, this may be more consequential than marginal differences among large pretrained backbones.
The result should nevertheless be interpreted conservatively. The present council comparison shows that homogeneous EVA-02, mixed EVA-02/EfficientNet-B5, and homogeneous EfficientNet-B5 councils all outperform single-teacher distillation, with the homogeneous EfficientNet-B5 council performing best numerically. This supports council distillation as a strong practical strategy, but it does not yet establish a complete theory of optimal teacher selection.

**3.4 Efficiency trade-offs**

The efficiency ranking (see Table 6) is not determined by parameter count or GMACs alone. EfficientNet-B0 was the smallest model by a large margin, with 4 million parameters and 0.38 GMACs, yet it was not the fastest model on the A100. ViT-S/16 DINO achieved the highest batch-64 throughput at 7381 images/s and the lowest peak memory use at 317.7 MB, despite requiring 4.2 GMACs. EfficientNet-B0 followed at 5956 images/s and 430.9 MB. The larger transformer models also mapped efficiently to the GPU: ViT-B/16 reached 3490 images/s despite 85.8M parameters and 16.9 GMACs. By contrast, EfficientNet-B5 had substantially lower throughput, reaching 1671 images/s despite requiring only 2.4 GMACs.
These results show that FLOPs are an incomplete proxy for realised inference speed. This likely reflects differences in hardware utilisation: dense transformer operations can map efficiently to modern GPU matrix units, whereas depthwise convolutional operations in EfficientNet-style models may be less favourable for high-throughput A100 inference. Consequently, the compactness of EfficientNet-B0 should not be interpreted as universal GPU throughput superiority. Among the evaluated models, ViT-S/16 achieved both the highest batch-64 throughput and the lowest peak allocated inference memory.
The tokenisation comparison also showed a clear efficiency cost. ViT-S/8 and ViT-S/16 have the same parameter count, but reducing the patch size from 16×16 to 8×8 increased compute from 4.2 to 16.7 GMACs, increased peak memory from 317.7 to 760.7 MB, and reduced batch-64 throughput from 7381 to 2146 images/s. The performance advantage of finer tokenisation therefore comes with a substantial compute and memory penalty.

The distillation results should be interpreted against this hardware-dependent efficiency background. The council-distilled EfficientNet-B0 student retains the inference cost of the 4 million parameter EN-B0 architecture while exceeding all individual backbones in classification performance. This makes it a strong compact student from a parameter and FLOP perspective. However, on A100 batch inference, ViT-S/16 provides a stronger raw throughput and memory point. The practical deployment frontier is therefore regime-dependent: EN-B0 is attractive as a low-parameter, low-GMAC student architecture, whereas ViT-S/16 is attractive for high-throughput server-GPU inference.

Accordingly, parameter count and GMACs should be interpreted alongside realised throughput, memory use, latency, and predictive performance rather than as sufficient measures of deployment efficiency.

### 3.5 Limitations and future work

The benchmark is restricted to a single public dataset acquired under controlled and standardised conditions. LIVECell crops are cleanly segmented, background-reduced, and morphologically consistent in ways that real laboratory data rarely are. The conclusions therefore apply most directly to controlled cropped phase-contrast classification. Performance under cross-instrument, cross-laboratory, or cross-modality conditions remains uncharacterised, and the relative architecture ordering may shift under stronger distribution shift.

The current evaluation uses single-cell crops in isolation. This suppresses field-level acquisition and contextual cues, allowing classification from intrinsic cellular morphology to be evaluated more directly, but it also removes information that may be useful in deployment. Real laboratory pipelines should integrate segmentation, tracking, quality control, and contextual information from neighbouring cells, local confluence, and colony organisation. A natural extension would combine a high-resolution crop of the target cell with a second, target-conditioned representation of its neighbourhood or the full field of view. The local and contextual branches could initially be combined as a teacher council or through learned feature fusion, and subsequently distilled into a compact deployment model. Evaluation on mixed-cell fields, together with context-masking and cross-experiment validation, would be required to determine whether any improvement reflects genuine cellular context rather than acquisition-specific shortcuts.

Ongoing work with in-house cultured cells across multiple microscopes, imaging modalities, and cell lines will address robustness and contextual classification more directly.

The LLRD result is similarly specific. We tested the original EVA-02 decay value as a recipe-transfer experiment and found that it degraded performance, while milder decay was near-neutral. This does not imply that all layer-wise learning-rate decay is harmful in microscopy, nor that the optimal fine-tuning recipe has been found. Stronger augmentation, alternative optimisers, different learning rates, in-domain self-supervised pretraining, and microscopy-specific regularisation may all shift the ranking.

Finally, all efficiency measurements were obtained on A100 GPUs under the present software and batching conditions. Relative throughput, latency, memory use, and deployment practicality may differ on consumer GPUs, CPUs, embedded systems, or microscope-integrated acquisition hardware. Peak allocated inference memory was measured under the stated benchmark configuration, but training memory and end-to-end pipeline latency, including image loading, preprocessing, and host-to-device transfer, were not systematically profiled. Future work should therefore evaluate accuracy, throughput, latency, memory, and robustness under realistic laboratory deployment conditions.
Taken together, these results motivate treating pretraining, fine-tuning, tokenisation, and distillation as first-order experimental variables rather than background implementation details. Biomedical imaging benchmarks can produce apparently decisive architecture conclusions when split structure, initialisation regime, or transfer protocol are left uncontrolled; once these are made explicit, the experimental regime in which models are trained and evaluated becomes more informative than the CNN-versus-ViT dichotomy itself. For practical label-free single-cell classification, the operative question is therefore not which family wins in the abstract, but which combination of architecture, pretraining, adaptation, and distillation gives the best accuracy-efficiency trade-off under a leakage-controlled protocol. Robustness under distribution shift remains a separate requirement for deployment.

# 4. Methods:

**4.1 Dataset Preparation**

This study utilised the LIVECell (Label-free In Vitro image examples of Cell) dataset, a large-scale resource for label-free phase-contrast microscopy (Edlund et al., 2021). The dataset consists of 5,239 images containing approximately 1.6 million annotated cells. It encompasses eight distinct cell lines representing a diverse range of morphologies: A172, BT-474, BV-2, Huh7, MCF7, SH-SY5Y, SkBr3, and SK-OV-3. As a highly normalised dataset with little background, it should help to provide a more robust foundation against artefacts that can be used by models for spurious correlations/shortcut learning.
LIVECell is originally annotated for instance segmentation rather than cell-line classification. We therefore converted the segmentation annotations into a single-cell classification dataset by extracting one crop per annotated cell instance. The resulting dataset was unbalanced, but this was not changed to test the effects of rare classes. Instances per class can be seen in Table 7. Each crop retained its class label and parent source-image identifier. Because multiple cells from the same microscopy field share acquisition conditions, illumination, focus, confluence, and local background statistics, random crop-level splitting can induce leakage between train and test partitions. We therefore constructed source-image-disjoint train/validation/test splits: parent images were first partitioned into 70/15/15 train/validation/test sets within each cell line, and all derived single-cell crops inherited the split assignment of their parent image. No source image contributed crops to more than one partition. All reported results use the full training partition and a fixed validation/test split across models and seeds.
Single-cell crops were read from per-class HDF5 files storing flattened pixel arrays and their original shapes. Each sample was reshaped to its stored dimensions and converted to a PIL image before transformation and batching. All images were resized to 224x224 pixels. During training we applied random horizontal flips ($p = 0.5$) and random rotations (±5°). Validation and test sets used only deterministic resizing and normalisation. Images were converted to tensors and normalised channel-wise with mean = 0.5 and std = 0.5 (applied per channel).

We intentionally used restrained augmentation rather than aggressive modern ViT recipes such as RandAugment, MixUp, CutMix, or strong colour jitter. The aim was to preserve comparability with the prior benchmark and avoid introducing architecture-specific augmentation advantages that would make the comparison less directly apples-to-apples on small phase-contrast single-cell crops. On this modality, stronger augmentation may or may not help, but testing that question lies outside the present study design.

**Table 7:** Instances per class and percentage of total dataset.

| Cell Type | Instances | Percentage |
|---|---|---|
| **A172** | 98,863 | 8 |
| **BT474** | 97,426 | 8 |
| **BV2** | 286,058 | 24 |
| **Huh7** | 26,350 | 2 |
| **MCF 7** | 251,940 | 21 |
| **SHSY5Y** | 193,596 | 16 |
| **SKOV3** | 46,894 | 3 |
| **SkBr3** | 199,059 | 16 |

### 4.2 Neural Architectures

We evaluated a focused set of convolutional and transformer-family models selected to test the main causal factors in the study. EfficientNet-B0 was used as a compact convolutional baseline and as the distillation student. EfficientNet-B5 represented a stronger mid-scale CNN. ViT-B/16 was used as the standard Vision Transformer baseline. ViT-S/8 and ViT-S/16 were included to test whether finer patch tokenisation benefits small single-cell crops. EVA-02 was included as a modern pretrained ViT-family model trained with masked image modelling with a CLIP feature-reconstruction target. All models were loaded from timm. For more detailed pretraining regimes see Table 8.

**Table 8:** Neural architectures and pretraining regimes.

| Model in paper | timm identifier used | Family | Pretraining |
|---|---|---|---|
| **EfficientNet-B0** | efficientnet_b0 | CNN / EfficientNet | ImageNet-1k |
| **EfficientNet-B5** | efficientnet_b5 | CNN / EfficientNet | ImageNet-1k |
| **ViT-B/16** | vit_base_patch16_224 | Vision Transformer | ImageNet-1k |
| **ViT-S/16 DINO** | vit_small_patch16_224.dino | Vision Transformer | SSL DINO on ImageNet-1k |
| **ViT-S/8 DINO** | vit_small_patch8_224.dino | Vision Transformer | SSL DINO on ImageNet-1k |
| **EVA-02-B/14** | eva02_base_patch14_224.mim_in22k | Vision Transformer | MIM on ImageNet-22k using EVA-CLIP |

### 4.3 Training Protocol

All models were trained in PyTorch using timm on NVIDIA A100 GPUs on ArchieWest. Optimisation used AdamW with learning rate 1×10-4, effective batch size 128 via gradient accumulation, restrained geometric augmentation, mixed precision training with bfloat16 where supported, and activation checkpointing for memory-intensive transformer models.

Pretrained models were fine-tuned end-to-end for up to 50 epochs with early stopping patience 5. Scratch-trained/randomly initialised models were trained for up to 200 epochs with patience 15 to avoid disadvantaging randomly initialised models through premature stopping. Since they were given substantially longer training budgets and earlier-stopping leniency, the pretraining advantage we report is conservative. Model selection used the checkpoint with lowest validation loss. To quantify training variability and ensure reproducibility, experiments were repeated across three seeds 42, 43, and 44, with deterministic seed control for Python, NumPy, PyTorch, and cuDNN.

The main pretrained benchmark used uniform fine-tuning unless otherwise specified. To test whether natural-image fine-tuning recipes transfer to phase-contrast microscopy, EVA-02 was additionally fine-tuned with layer-wise learning-rate decay factor 0.70, corresponding to the original EVA-02 fine-tuning recipe and therefore treated as a recipe-transfer test rather than as an arbitrary hyperparameter sweep.

### 4.4 Matched pretraining comparison

To isolate pretraining from architecture family, EfficientNet-B5 and ViT-B/16 were evaluated under both pretrained and random initialisation using matched data splits, augmentation, optimiser, effective batch size, and evaluation protocol. This within-architecture comparison quantifies how much of the observed CNN-ViT ranking is attributable to initialisation rather than architecture alone.

### 4.5 Knowledge Distillation

Knowledge distillation experiments use a combined hard-label and soft-label objective. Student logits were trained against ground-truth labels via cross-entropy (hard loss) and against temperature-scaled teacher softmax outputs via KL divergence (soft loss), with the total loss weighted as:

$$\mathcal{L} = \alpha\mathcal{L}_{hard} + (1-\alpha)T^2\mathcal{L}_{soft} \tag{1}$$

where temperature $T \in \{3, 4\}$ and $\alpha = 0.2$, placing 80% of the loss weight on the soft targets.

We evaluated both single-teacher and council-based distillation. In the single-teacher setting, soft targets were provided by a single pretrained teacher, with EVA-02 → EfficientNet-B0 and EfficientNet-B5 → EfficientNet-B0 as the principal pairings. In the council setting, the teacher target was constructed by averaging the logits of K teachers sample-wise,

$$\boldsymbol{z}_{council} = \frac{1}{K}\sum_{k=1}^{K}\boldsymbol{z}^{(k)} \tag{2}$$

after which the averaged council logits were converted into a temperature-scaled soft target distribution via:

$$\boldsymbol{p}_{council} = softmax\left(\frac{\boldsymbol{z}_{council}}{T}\right) \quad (3)$$

We compared a homogeneous EVA-02×3 council against a mixed 2×EVA-02 + EfficientNet-B5 council and a homogeneous EfficientNet-B5x3 council, while holding the student architecture fixed. All distillation runs used the same student training protocol as the standard fine-tuning experiments, including pretrained student initialisation, optimiser, augmentations, data fractions, validation procedure, and evaluation pipeline, so that differences could be attributed to teacher identity or council composition rather than to changes in student training. All distillation runs furthermore used the same corrected source-image-disjoint splits as the main benchmark.

### 4.6 Evaluation Metrics

Classification performance was assessed using test accuracy, macro-F1, balanced accuracy, and expected calibration error.

Model complexity was computed uniformly with *PyTorch torch.utils.flop_counter.FlopCounterMode* using a synthetic 1×3×224×224 input in evaluation mode. We report GMACs as the returned forward FLOP count divided by two, so that one multiply-add is treated as one MAC. The raw FLOP count, PyTorch version, timm version, model name, and input shape were stored with the profiling output for reproducibility.

Inference profiling used synthetic tensors rather than the dataloader as well as torch.inference_mode(), bf16 autocast, CUDA event timing, fixed 224×224 inputs, cuDNN benchmark mode enabled, and a single A100SXM GPU. Peak memory denotes CUDA peak allocated memory during inference, not training memory

Accuracy-compute trade-offs were visualised by plotting test balanced accuracy and macro-F1 against GMACs, with compact distilled models evaluated by student inference cost only.

**Data availability.** The LIVECell dataset is publicly available (Edlund et al., 2021). The single-cell crops used in this study are deposited on the University of Strathclyde Pure portal (DOI: 10.15129/f9a95003-e23d-4a29-bf2e-7451ccc563ae, CC BY 4.0). Source-image-disjoint split assignments (parent-image identifiers per partition) will be released at https://github.com/Smart-Miscroscopy-Lab.

**Code availability.** Training, distillation, and profiling code, together with configuration files and trained-model checkpoints, will be made available at https://github.com/Smart-Miscroscopy-Lab.

**Acknowledgements.** Results were obtained using the ARCHIE-WeSt High Performance Computer (www.archie-west.ac.uk) based at the University of Strathclyde.